\documentclass[letterpaper]{article} 
\usepackage{aaai25}  

\usepackage{multirow}
\usepackage{subcaption}
\usepackage{array}
\usepackage{booktabs}
\usepackage{longtable}
\usepackage{xcolor}
\usepackage{times}  
\usepackage{helvet}  
\usepackage{courier}  
\usepackage[hyphens]{url}  
\usepackage{graphicx} 
\usepackage{natbib}  
\usepackage{caption} 
\usepackage{algorithm}
\usepackage{algorithmic}

\usepackage{newfloat}
\usepackage{listings}
\DeclareCaptionStyle{ruled}{labelfont=normalfont,labelsep=colon,strut=off} 
\floatstyle{ruled}
\newfloat{listing}{tb}{lst}{}
\floatname{listing}{Listing}
\title{FinTruthQA: A Benchmark for AI-Driven Financial Disclosure Quality Assessment in Investor–Firm Interactions}
\author{
    Peilin Zhou\textsuperscript{\rm 1}\equalcontrib,
    Ziyue Xu\textsuperscript{\rm 2}\equalcontrib,
    Xinyu Shi\textsuperscript{\rm 3},
    Jiageng Wu\textsuperscript{\rm 4},
    Yikang Jiang\textsuperscript{\rm 5},\\
    Dading Chong\textsuperscript{\rm 6},
    Wang Dong\textsuperscript{\rm 2},
    Jun Chen\textsuperscript{\rm 2},
    Bin Ke\textsuperscript{\rm 5},
    Jie Yang\textsuperscript{\rm 4}\thanks{Corresponding author. This work was conducted while the author was affiliated with Zhejiang University.}
}

\affiliations{
    \textsuperscript{\rm 1}New York University Abu Dhabi\\
    \textsuperscript{\rm 2}Zhejiang University\\
    \textsuperscript{\rm 3}Southwest University of Finance and Economics\\
    \textsuperscript{\rm 4}Harvard University\\
    \textsuperscript{\rm 5}National University of Singapore\\
    \textsuperscript{\rm 6}Peking University\\
    \{zhoupalin, zyxu1224, jieynlp\}@gmail.com
}

\begin{document}

\maketitle

\begin{abstract}
Accurate and transparent financial information disclosure is essential for market efficiency, investor decision-making, and corporate governance. Chinese stock exchanges' investor interactive platforms provide a widely used channel through which listed firms respond to investor concerns, yet these responses are often limited or non-substantive, making disclosure quality difficult to assess at scale. To address this challenge, we introduce FinTruthQA, to our knowledge the first benchmark for AI-driven assessment of financial disclosure quality in investor--firm interactions. FinTruthQA comprises 6,000 real-world financial Q\&A entries, each manually annotated based on four key evaluation criteria: \textit{question identification}, \textit{question relevance}, \textit{answer readability}, and \textit{answer relevance}. We benchmark statistical machine learning models, pre-trained language models and their fine-tuned variants, as well as large language models (LLMs), on FinTruthQA. Experiments show that existing models achieve strong performance on \textit{question identification} and \textit{question relevance} (F1\textgreater 95\%), but remain substantially weaker on \textit{answer readability} (Micro F1$\sim$88\%) and especially \textit{answer relevance} (Micro F1$\sim$80\%), highlighting the nontrivial difficulty of fine-grained disclosure quality assessment. Domain- and task-adapted pre-trained language models consistently outperform general-purpose models and LLM-based prompting on the most challenging settings. These findings position FinTruthQA as a practical foundation for AI-driven disclosure monitoring in capital markets, with value for regulatory oversight, investor protection, and disclosure governance in real-world financial settings. The dataset, code, annotation guidelines, hyperparameters, and extended error analyses are publicly available at \url{https://github.com/bethxx99/FinTruthQA}.
\end{abstract}

\section{Introduction}

In capital markets, listed firms increasingly communicate with investors through online interactive platforms, where the quality of their responses can shape investor confidence, disclosure governance, and market efficiency. The online Q\&A platforms established by the Shanghai Stock Exchange (SSE) and Shenzhen Stock Exchange (SZSE) provide a widely used channel through which listed firms respond to investor concerns in near real time. Unlike conventional disclosures (e.g., mandatory reports or conference calls), which follow relatively standardized formats, these platforms capture more spontaneous and fine-grained investor--firm interactions, making them a valuable setting for evaluating how effectively firms communicate substantive information to the market.

Despite their practical importance, these platforms present substantial challenges for large-scale analysis because investor--firm interactions are both massive in volume and highly variable in quality. Traditional disclosure measures primarily capture quantity rather than quality~\citep{leuz2016economics}. Although natural language processing (NLP) has successfully automated the analysis of financial reports~\citep{smailovic2017automatic} and news sentiment~\citep{van2015fine}, there is currently no standard benchmark for systematically evaluating AI models on financial disclosure quality in real-world investor--firm interactions, especially for tasks requiring contextual and relevance-sensitive judgment.

To address this gap, we present FinTruthQA, a benchmark dataset for AI-driven financial disclosure quality assessment on Chinese stock exchanges' interactive investor platforms. It is organized around four key evaluation criteria, namely \textit{question identification}, \textit{question relevance}, \textit{answer readability}, and \textit{answer relevance}, which together capture important dimensions of disclosure quality in investor--firm interactions and define the four benchmark tasks in our study. Using FinTruthQA, we systematically evaluate a broad range of NLP techniques across these tasks. Our results support data-driven disclosure monitoring, regulatory oversight, and investor protection in real-world financial settings.

\begin{figure*}[t]
\begin{center}
\resizebox{\textwidth}{!}{
  \includegraphics{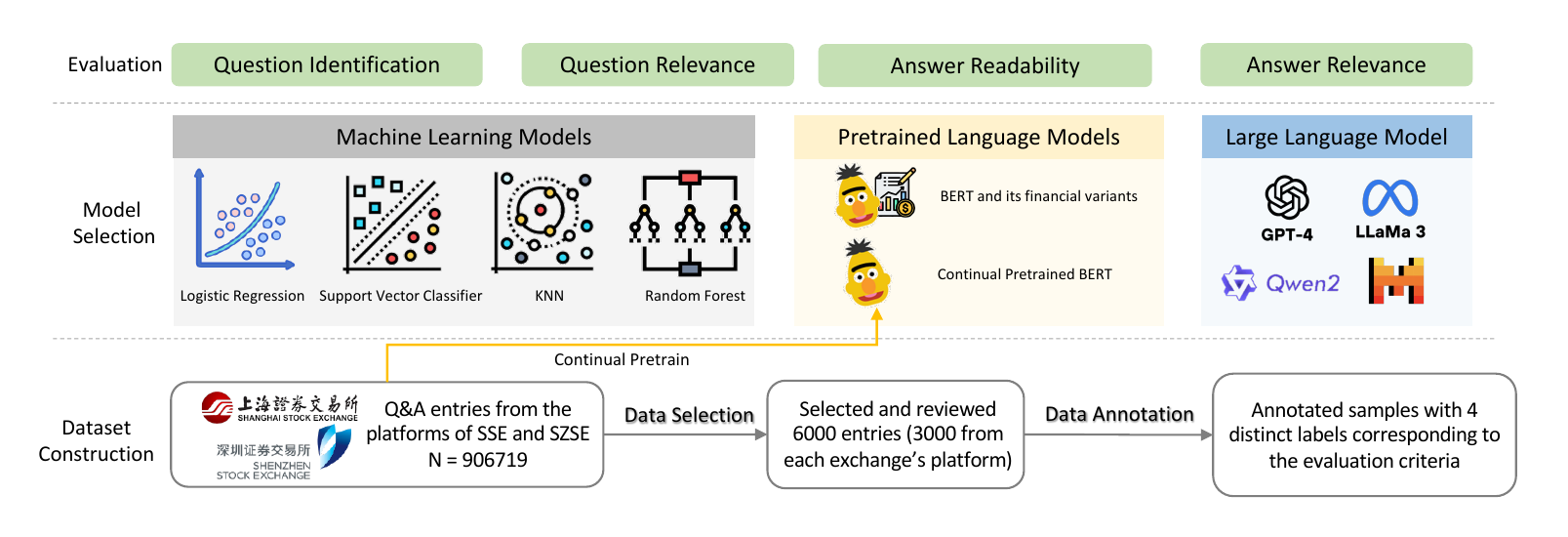}
}
\end{center}
\caption{Overview of the FinTruthQA construction and evaluation pipeline. The process begins with dataset construction, where 906,719 Q\&A entries were collected from the official platforms of the Shanghai Stock Exchange (SSE) and Shenzhen Stock Exchange (SZSE). From these, a subset of 6,000 entries (3,000 per platform) was selected and manually annotated based on four evaluation criteria. For model evaluation, we include traditional machine learning models, pre-trained language models, and large language models (LLMs), enabling comparative analysis of performance across different modelling strategies.}
\label{fig:overview}
\end{figure*}

Using FinTruthQA, we obtain the following key findings:
\begin{itemize}
    \item Among the four tasks, \textit{answer readability} and especially \textit{answer relevance} are substantially more difficult than \textit{question identification} and \textit{question relevance}, highlighting the nontrivial nature of fine-grained disclosure quality assessment.
    \item On the most challenging task, \textit{answer relevance}, financial-domain PLMs, particularly FinBERT with continued pretraining, achieve the best performance, reaching a Micro F1 of 80.04\%.
    \item Current general-purpose LLMs still underperform domain-adapted supervised models on fine-grained relevance judgment, indicating that stronger domain grounding is needed for trustworthy deployment in financial applications.
\end{itemize}

\section{Related work}

\subsection{Disclosure Measurement} Measuring disclosure quality has been a persistent challenge in empirical accounting research~\citep{leuz2016economics}. Comprehensive measures traditionally quantify disclosure frequency, such as counting reports or conference calls~\citep{lee2022shall}. Other approaches rely on analyst surveys—which are often limited to specific, large U.S. firms~\citep{nagar2003discretionary}—or utilize keyword-based techniques to analyze the tone of traditional, structured disclosures like earnings forecasts~\citep{bochkay2023textual,frankel2010pennies}. Crucially, these traditional methods primarily capture information quantity or basic sentiment, making it difficult to evaluate the substantive quality of dynamic, unstructured text.

\subsection{Financial NLP} Due to the technical complexity of financial text~\citep{chang2016measuring}, extensive research has developed domain-adapted pre-trained models~\citep{yang2020finBERT,liu2021finBERT}. Early applications targeted sentiment analysis~\citep{wan2021sentiment}, distress prediction~\citep{hajek2024corporate}, and named entity recognition~\citep{zhang2024chinese}. Recently, the focus has shifted toward question-answering (QA) involving textual and tabular numerical reasoning~\citep{zhao2022multihiertt,li2022finmath}, alongside the deployment of large language models (LLMs). Several domain-specific LLMs—such as FinTral~\citep{bhatia2024fintral}, FinBERT2~\citep{xu2025finbert2}, and Open‑FinLLMs~\citep{xie2024open}—have demonstrated robust capabilities across multimodal and text-based financial tasks. Concurrently, new datasets and benchmarks like CFBenchmark~\citep{lei2023cfbenchmark} and FinBen~\citep{xie2024finben} evaluate LLM performance on financial assistance, corporate QA, and data annotation~\citep{aguda2024large}. However, despite these advancements, no prior dataset is dedicated to the automated quality assessment of financial Q\&A interactions.
\begin{figure*}[t]
\begin{center}
\includegraphics[width=\textwidth]{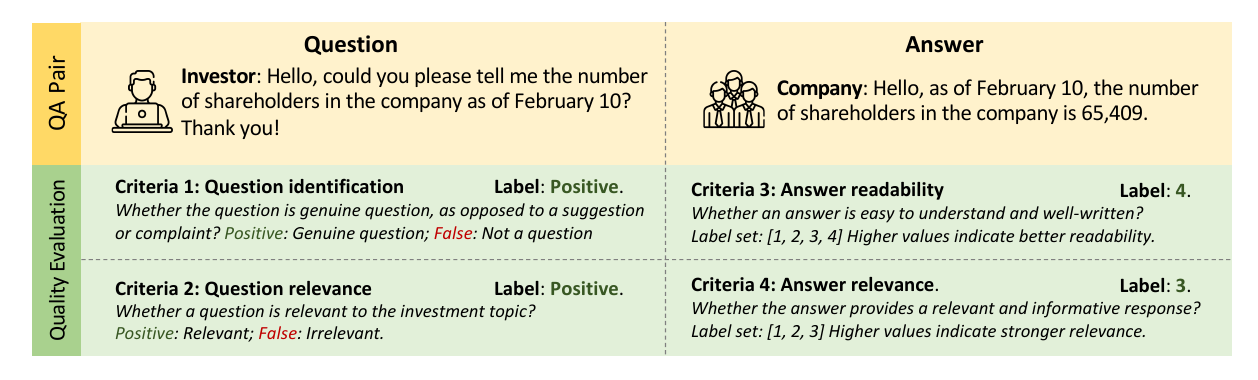}
\end{center}
\caption{Example of a Q\&A interaction between a retail investor and a listed company representative drawn from the FinTruthQA dataset (translated to English), along with the annotated labels for each of the four quality assessment criteria: question identification (whether the input is a genuine question), question relevance (whether the question is related to investment-related topics), answer readability (how clear and well-written the answer is, on a 1–4 scale), and answer relevance (how well the answer addresses the question, on a 1–3 scale).}
\label{fig:eg}
\end{figure*}
\section{Dataset Construction}


\subsection{Data Collection}
The data for this study was obtained from the interactive platforms for communication between investors and listed companies established by the Shanghai Stock Exchange (Shangzhen E-Hudong \url{https://sns.sseinfo.com/}) and the Shenzhen Stock Exchange (Hudongyi \url{https://irm.cninfo.com.cn/}). These two stock exchange sites cover almost all listed companies in China. These Q\&A interactions are authentic, generated by human users and company representatives, and are regulated by the China Securities Regulatory Commission, ensuring the accuracy of disclosures. The companies involved are legally obligated to ensure the accuracy of their disclosures, and failure to do so could result in legal consequences. This regulatory oversight ensures that the data we used reflects genuine interactions. We collected over 900,000 Q\&A entries, among which we randomly selected 6,000 samples for manual annotation, with 3,000 entries from each exchange's platform. The data collection involved using web scraping techniques to extract the Q\&A entries from the official platforms. The data were then manually reviewed and processed to ensure quality and consistency.  The process was overseen by a PhD student from the Business School of National University of Singapore. The ethics statement, including considerations of general ethical conduct and potential negative societal impacts, can be found in our GitHub repository. We also give several examples in our GitHub repository to demonstrate the wide range of investor inquiries covered in our dataset.

\subsection{Data Annotation}

We focus on four key information disclosure quality evaluation criteria\footnote{In this paper, the terms "criteria" and "tasks" are used interchangeably. "Criteria" refers to the specific standards used to evaluate the quality of the Q\&A pairs, while "tasks" refers to the individual evaluation processes performed based on these criteria, particularly in the experimental sections.}: \textit{question identification}, \textit{question relevance}, \textit{answer relevance}, and \textit{answer readability}. These criteria are widely recognized as crucial indicators of information quality and are important for investors to consider when evaluating Q\&A information. Specifically, the criteria in FinTruthQA are defined as follows:

\begin{enumerate}
\item \textbf{Question identification}: This criterion assesses whether the question from an investor is a genuine question, as opposed to a suggestion or complaint. Each question is assigned a binary label: \textit{Positive}, indicating it is a real question, or \textit{Negative}, indicating it is not. Data entries with questions labeled as \textit{Negative} are excluded from further evaluation under the remaining criteria.

\item \textbf{Question relevance}: This criterion evaluates whether a question is pertinent to the investment topic, labeling it as \textit{Positive} if relevant and \textit{Negative} if not. It assesses whether the question is related to the company's shareholders, financial indicators, industry conditions, or other investment-related topics.

\item \textbf{Answer readability}: This criterion evaluates how easily an answer can be understood and whether it is well-written. It considers multiple dimensions, including structural coherence, clarity of information, stylistic appropriateness, and grammatical correctness, ensuring a comprehensive evaluation of the answer's quality. The readability score ranges from 1 to 4, with 4 indicating the highest level of readability—clear, concise, and free of ambiguity. A score of 3 indicates moderate readability with some issues, while a score of 2 reflects significant readability challenges. A score of 1 suggests the answer is unreadable, with major ambiguities, incoherent sentences, or content unrelated to the question.

\item \textbf{Answer relevance}: This criterion evaluates whether an answer is relevant and informative in response to the question. It assesses whether the answer provides sufficient detail and explanation to address the question. Answers are classified into three levels: Level 1 indicates the answer is completely unrelated to the question (none of the questions is answered); Level 2 indicates the answer is partially related (some questions are answered or it is difficult to judge) ; and Level 3 indicates the answer fully responds to the question (all questions are answered).

\end{enumerate}

The importance of these criteria lies in their collective ability to enhance the quality and reliability of information on investor communication platforms. The annotation process was carried out in three phases: pre-annotation, formal annotation, and data verification and integration. In the pre-annotation phase, we recruited six students from the Accounting Department at Zhejiang University and conducted five rounds of annotation. After the last round, the inter-annotator agreement (IAA) rate was calculated, and the four students with the highest average IAA rates (85.52\%)—two undergraduates and two master’s students—were selected for the formal annotation phase.
To ensure annotation quality, the four selected annotators were evenly divided into two groups, with each group annotating the same data instances. Discrepancies were first discussed and resolved within the group, with unresolved issues escalated to the project leader for final decisions. After the formal annotation, the project leader verified the data to ensure alignment with the latest annotation criteria and completeness, before integrating the data and summarising key information. Figure~\ref{fig:eg} illustrates an example of the annotation. A detailed description of the
annotation process is listed in our GitHub repository.

\subsection{Quality Control}
For the \textit{answer relevance} task, we observed that the raw annotations contained noise due to the subjective nature of this task. To address this, we introduced an additional quality control step to correct the raw annotations. We implemented an enhanced correction process inspired by Confident Learning principles~\citep{northcutt2021confident}. Sample data points demonstrating inaccuracies, along with implementation details, can be found in our GitHub repository.

\subsection{Dataset Statistics}
\label{stats}
\begin{figure}[t]
  \centering
      \includegraphics[width=0.43\linewidth]{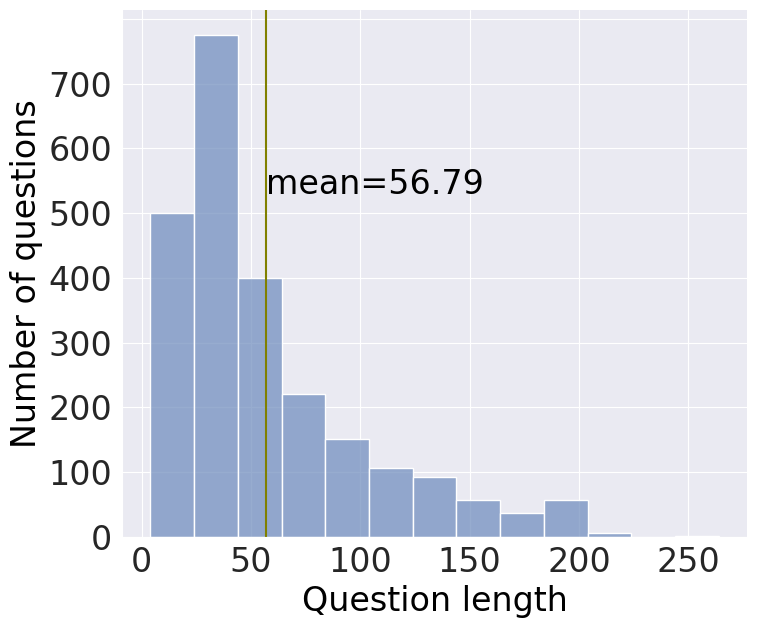} 
      \quad\quad\quad
      \includegraphics[width=0.43\linewidth]{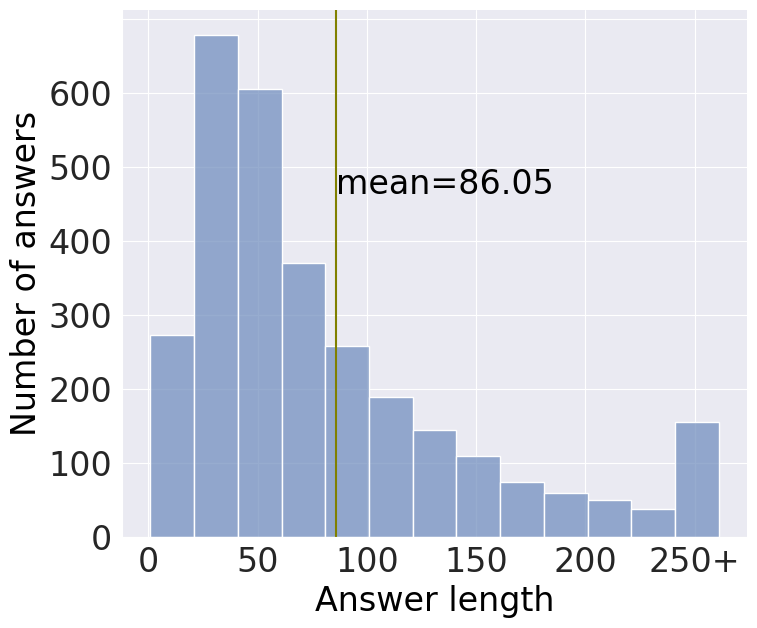}
  \caption{Length distributions of our dataset (in characters).}
  \label{fig:dist}
\end{figure}

The complete annotated dataset comprises 6,000 samples labeled for \textit{question identification}, \textit{question relevance}, \textit{answer relevance}, and \textit{answer readability} tasks, with each data point having four distinct labels corresponding to these tasks. The dataset was randomly divided into training, validation, and testing sets in a ratio of 4:1:1. The label distributions for each task are detailed in Table~\ref{dist}. The provided statistics reflect the data after undergoing the quality control process. Note that our later analysis focuses exclusively on real questions. Only for the first task (\textit{question identification}), we used the full set of 6,000 samples. For the remaining three tasks, we removed non-question samples, as they could skew the model's understanding and performance, especially for tasks like \textit{question relevance}, where the context of a real question is crucial for accurate assessment. It is worth noticing that the \textit{question relevance} labels are highly imbalanced, with only 6 and 7 negative labels in the validation and test sets, respectively. We retained this task as the imbalance reflects the real-world data distribution, and the concept of \textit{question relevance} is important in financial information disclosure.  Figure~\ref{fig:dist} shows the length distribution of questions and answers, measured in characters, for all 6,000 QA pairs.

\begin{table}[t]
\centering
\resizebox{1\linewidth}{!}{
\begin{tabular}{cccccccc}
\toprule
\textbf{Task} & \textbf{Label} & \textbf{\# Train} & \textbf{\# Valid} & \textbf{\# Test} & \textbf{\# Label} & \textbf{Freq.} & \textbf{Total} \\
\midrule
\multirow{2}{*}{Question identification} & Negative & 327  & 90  & 81  & 498  & 8.30\%  & \multirow{2}{*}{6,000} \\
                        & Positive & 3,673 & 910 & 919 & 5,502 & 91.70\% &                       \\
                        \midrule
\multirow{2}{*}{Question relevance}      & Negative & 714  & 6   & 7   & 27   & 0.49\%  & \multirow{2}{*}{5,502} \\
                        & Positive & 3,654 & 911 & 910 & 5,475 & 99.51\% &                       \\
                        \midrule
\multirow{4}{*}{Answer readability}      & 1 & 106  & 190 & 198 & 1,094 & 19.88\% & \multirow{4}{*}{5,502} \\
                        & 2 & 26   & 3   & 6   & 35   & 0.64\%  &                       \\
                        & 3 & 68   & 19  & 20  & 107  & 1.94\%  &                       \\
                        & 4 & 2,868 & 705 & 693 & 4,266 & 77.54\% &                       \\
                        \midrule
\multirow{3}{*}{Answer relevance}        & 1 & 731  & 194 & 206 & 1,131 & 20.63\% & \multirow{3}{*}{5502} \\
                        & 2 & 533  & 136 & 130 & 779  & 14.21\% &                       \\
                        & 3 & 2,404 & 587 & 581 & 3,572 & 65.16\% &                       \\
                        \bottomrule
\end{tabular}
}

\caption{Label distribution and dataset statistics for the four quality evaluation tasks in FinTruthQA. Note that the total number of labeled instances for the latter three tasks excludes entries identified as non-questions during the question identification stage.}

\label{dist}
\end{table}

\begin{table*}

  \centering
  \resizebox{1\linewidth}{!}{
  \begin{tabular}{llcccccccc}
    \toprule
    \multirow{2}*{\textbf{Category}} & \multirow{2}*{\textbf{Model}}&\multicolumn{4}{c}{\textbf{Question identification (\%)}}& \multicolumn{4}{c}{\textbf{Question relevance(\%)}}                   \\
    \cmidrule(r){3-6} \cmidrule(r){7-10}
   & &  \textbf{Accuracy}   & \textbf{Precision} & \textbf{Recall}  &  \textbf{F1} &  \textbf{Accuracy}     & \textbf{Precision} & \textbf{Recall} &  \textbf{F1}\\
     \hline
     \multirow{4}*{ML-based}&TF-IDF + LR  &  93.70$\pm$0.86  &94.11$\pm$0.75 &99.35$\pm$0.18 & 96.66$\pm$0.48 &  99.42$\pm$0.13 &99.42$\pm$0.13 &\textbf{100.00}$\pm$0.00  & 99.71$\pm$0.07\\
     &TF-IDF + RF  &  93.70$\pm$0.65&94.02$\pm$0.63 &99.46$\pm$0.18 & 96.66$\pm$ 0.37&  99.42$\pm$0.13  &99.42$\pm$0.13 &99.93$\pm$0.10& 99.67$\pm$0.08\\
     &TF-IDF + SVM  &  93.60$\pm$0.73 &93.78$\pm$0.76 &\textbf{99.64}$\pm$0.13& 96.62$\pm$ 0.40&  99.42$\pm$0.13 &99.42$\pm$0.13 &\textbf{100.00}$\pm$0.00& 99.71$\pm$0.07\\
     &TF-IDF + KNN  &  92.67$\pm$1.11  &93.27$\pm$0.98&99.16$\pm$0.37& 96.12$\pm$0.61&  99.42$\pm$0.13 &99.42$\pm$0.13 &\textbf{100.00}$\pm$0.00 & 99.71$\pm$0.07\\
     
    \hline
    PLM-based&BERT &  \textbf{96.89}$\pm$0.55 &\textbf{97.64}$\pm$0.22&99.02$\pm$0.41& \textbf{98.32}$\pm$0.28&  \textbf{99.67}$\pm$0.11 &\textbf{99.72}$\pm$0.12&99.96$\pm$0.05& \textbf{99.85}$\pm$0.06\\ 
    \bottomrule
  \end{tabular}
}

  \caption{Model evaluation results of machine learning and pre-trained language models on the \textit{question identification} and \textit{question relevance} task. Models include TF-IDF-based classifiers (logistic regression, random forest, SVM, KNN) and a pre-trained BERT model. The PLM-based BERT achieves the highest overall performance across both tasks.}
  \label{eval12}

\end{table*}

\section{Benchmarks}

\subsection{Model Selection}
To fully evaluate the performance of different methods in FinTruthQA, we benchmarked several models that cover Machine Learning (ML)-based, Pre-trained Language Model (PLM)-based, PLM with continued Pre-Training (CPT)-based, and LLM-based approaches. 
Given the distinct nature and varying difficulty of the tasks, we applied tailored approaches to each accordingly.

\textbf{ML-based.} 
We modeled these tasks as text classification problems and used various advanced machine learning models, including Logistic Regression (LR), Support Vector Machine (SVM), K-nearest neighbor (KNN), and Random Forest (RF). 
For semantic feature extraction, we first utilized the Jieba toolkit \footnote{\url{https://github.com/fxsjy/jieba}} for Chinese word segmentation and then extracted weighted text representations using TF-IDF.
To achieve optimal performance, we used cross-validation and GridSearch to rigorously explore the best hyperparameters. The best-performing classifiers and hyperparameters varied across tasks. Therefore, we performed task-specific hyperparameter searches for each task individually.

\textbf{PLM-based.} 
In our experiments, we found that the tasks of \textit{question identification} and \textit{question relevance }were relatively straightforward, with BERT achieving excellent performance on these tasks.
Therefore, our focus shifted to exploring the effectiveness of PLM-based methods on the more challenging tasks of \textit{answer relevance} and \textit{answer readability}.
Specifically, we investigated a range of BERT-based models, including those 
specifically trained on financial text data, such as financial news, research reports, and financial encyclopedia entries (\textit{FinBERT,}\footnote{\url{https://github.com/valuesimplex/FinBERT}} \textit{Mengzi-fin}~\citep{zhang2021mengzi}, and \textit{DKPLM}~\citep{zhang2022dkplm}). 
Furthermore, we examined whether fine-tuned PLMs on specific tasks, such as NLI (Natural Language Inference) and QA (Question Answering), would offer any benefits for these tasks (\textit{SBERT-nli} ~\citep{reimers2019sentence} and \textit{RoBERTa-extractive-qa} ~\citep{liu2019roberta}). 
Finally, we also examined the performance of a large BERT model (\textit{BERT (Large)}).\footnote{\url{https://huggingface.co/yechen/bert-large-chinese}} 
Notably, unless otherwise specified, all PLMs employed in this study are the base and Chinese versions. Links for these PLMs can be found in our GitHub repository.

\textbf{PLM with CPT-based.} 
General domain PLMs, trained on broad texts, often lack the specialized knowledge required to handle the nuanced language and terminology in finance. This limitation is especially pronounced in tasks like \textit{answer relevance}, where understanding domain-specific jargon and abbreviations is essential for accurate predictions.
To bridge this gap, we curated a substantial volume of domain-specific texts and conducted continued pre-training on general PLMs. 
Specifically, we collected over 900,000 Q\&A pairs from the SSE and SZSE platforms by web scraping.  The top-performing models from previous experiments were then selected for continued pre-training using masked language modeling (MLM). 
This pre-training on a large-scale, domain-specific dataset enhances the PLM's ability to understand financial contexts, capture more specialized information, and achieve better performance in downstream tasks.

\textbf{LLM-based.} 
We further evaluate a diverse set of advanced LLMs on the most challenging \textit{answer relevance} task, including the OpenAI series (GPT-4~\citep{achiam2023gpt}, GPT-4o, GPT-5, and O3), GPT-OSS 120B, LLaMA-3.1~\citep{dubey2024llama}, the Qwen series (Qwen2~\citep{yang2024qwen2} and Qwen3-Next variants), and Mistral~\citep{jiang2023mistral}. Due to their in-context learning ability, LLMs can perform this task in a zero-shot manner, without requiring specific fine-tuning. Specifically, we provided the LLMs with task descriptions and evaluation criteria, instructing them to directly output a relevance score (\textit{Direct}). Moreover, we incorporated the Chain of Thought (CoT)~\citep{cot-first-wei2022} technique, prompting the models to provide specific reasoning alongside the score (\textit{CoT}). Full details are provided in our GitHub repository.

\subsection{Settings and Metrics}
\textbf{Implementation Details.} For the ML-based models, we utilized the standard \texttt{scikit-learn}\footnote{\url{https://scikit-learn.org/stable/}} library. For all PLM-based methods, we used \texttt{PyTorch}\footnote{\url{https://github.com/pytorch/pytorch}} and HuggingFace’s framework\footnote{\url{https://github.com/HuggingFace}} for downstream fine-tuning and evaluation. Detailed hyperparameter settings for our PLM-based experiments can be found in our GitHub repository.
We conducted experiments comparing Q\&A pairs with and without the company name as a prefix and found negligible differences, so we ultimately used data without the prefix. The tasks of \textit{question identification} and \textit{question relevance} only required encoding the questions as input, while \textit{answer readability} and \textit{answer relevance} required encoding both the question and the answer. The sequence started with a special classification token [CLS], and the question and answer were separated by the [SEP] token . 
During continued pre-training, we added two special tokens, [Q] and [A], into the vocabulary and prefixed them to the question and answer, respectively. This helps the PLMs better identify the components of each sentence. The continued pre-training was performed based on the UER framework~\citep{zhao2019uer}. All tests of LLMs were conducted using the transformer framework\footnote{\url{https://github.com/huggingface/transformers}}, except for the GPT-4 experiments, which were conducted by calling OpenAI's API with version GPT-4-0613. Specifically, we used Llama-3.1 with 8 billion parameters, and both Qwen2 and Mistral with 7 billion parameters.

\paragraph{Metrics.}
For \textit{question identification} and \textit{question relevance}, which are binary classification tasks, we used accuracy, precision, recall, and F1-score as evaluation metrics. 
For \textit{answer readability} and \textit{answer relevance}, which are multi-class classification tasks, we calculated both micro and macro F1-scores, along with the Quadratic Weighted Kappa (QWK)~\citep{dong2017attention}, which is well-suited for ordinal classification by penalizing misclassifications based on their distance from the true label. Each model was run in five times, and the mean and variance of the results are reported.

\subsection{Results and Analysis}
In this section, we present the evaluation results and discuss the performance of the benchmarked models. To provide a comprehensive understanding, our analysis is organized systematically across the four aforementioned tasks, highlighting the strengths and limitations of different modeling approaches.
\subsubsection{Task 1: Question Identification}

\begin{table}[t]

  \centering
  \resizebox{1\linewidth}{!}{
  \begin{tabular}{l|lcccc}
    \toprule
   \textbf{Category} & \textbf{Model} &  \textbf{F1 (Mi)}    &  \textbf{F1 (Ma)}  & \textbf{F1 (W)} & \textbf{QWK}\\
   
     \hline
     \multirow{4}*{ML-based}&TF-IDF + LR  &  \textbf{81.72} $\pm$0.27&\textbf{34.91}$\pm$0.22&\textbf{78.90}$\pm$0.27 & \textbf{41.20}$\pm$0.80\\
     
     &TF-IDF + RF  &  80.48$\pm$0.70 & 32.81$\pm$1.06&76.85$\pm$1.03 &33.79$\pm$3.87\\
     
     &TF-IDF + SVM  &  81.64$\pm$0.37 & 33.92$\pm$0.38 &78.15$\pm$ 0.40
     &37.98$\pm$1.48\\
     
     &TF-IDF + KNN  &  78.99$\pm$0.34 & 31.07$\pm$1.23 &74.95 $\pm$0.84&27.31$\pm$4.28\\
     
    \hline
    \multirow{8}*{PLM-based}&BERT & 86.04$\pm$0.44 &41.97$\pm$0.67 &84.89$\pm$0.44 &63.52$\pm$1.43\\ 
    
    

    &RoBERTa & 86.26$\pm$0.62 & 42.02$\pm$1.92 & 85.08$\pm$0.89 & 64.00$\pm$0.96\\
    
    &FinBERT &  \textbf{87.57}$\pm$0.39 &41.34$\pm$0.45 &\textbf{85.98}$\pm$0.69 &\textbf{66.53}$\pm$1.83\\
    
    &Mengzi-fin & 86.99$\pm$0.80&41.93$\pm$1.42&85.77$\pm$1.20 &66.33$\pm$1.81\\
    
    &DKPLM  &  87.24$\pm$0.76 & 41.90$\pm$1.17 & 85.76$\pm$0.72 & 65.78$\pm$0.63\\ 
    
    &SBERT-nli &  85.31$\pm$1.05 & 39.28$\pm$0.81 & 83.48$\pm$1.38 & 58.19$\pm$2.77\\ 
    
    &RoBERTa-extractive-qa & 87.21$\pm$0.52  & 41.22$\pm$0.11 & 85.75$\pm$0.64 & 65.97$\pm$0.80\\ 

    &BERT (Large) & 87.24$\pm$1.03  & 42.17$\pm$1.66 & 85.78$\pm$1.30 & 65.92$\pm$2.74\\  

    &RoBERTa (Large) & 86.48$\pm$1.48&\textbf{45.51}$\pm$4.59&85.37$\pm$1.41 &64.58$\pm$2.11\\ 
    
    \bottomrule
  \end{tabular}
  }

\caption{Model evaluation results on the \textit{answer readability} task. 
PLM-based models consistently outperform ML-based baselines across all metrics, with FinBERT achieving the highest F1 (micro and weighted) and QWK, indicating strong alignment with human readability judgments.
}
\label{eval3}
\end{table}

In this task, we benchmarked several ML-based and PLM-based models, with their performances summarized in Table~\ref{eval12}. The ML-based models showed consistent results with minor variations. Both TF-IDF + LR and TF-IDF + RF reached an accuracy of 93.70\% and an F1 score of 96.66, with closely aligned precision and recall. TF-IDF + SVM exhibited slightly lower accuracy but achieved the highest recall among the ML models, demonstrating its strength in identifying relevant instances. TF-IDF + KNN, while performing well with an accuracy of 92.67\%, had the lowest precision and F1 score in this group.
On the other hand, the PLM-based BERT model outperformed all ML-based methods, achieving the highest accuracy, precision, and F1 score. This result highlights the PLM model’s ability to capture deeper contextual nuances, which traditional frequency-based methods like TF-IDF struggle to achieve.

\subsubsection{Task 2: Question Relevance}

Table~\ref{eval12} presents the evaluation results for the question relevance task. The PLM-based model outperformed the ML-based models on all metrics except recall, achieving an accuracy of 99.67\%, an F1 score of 99.85\%, precision of 99.72\%, and recall of 99.96\%. This demonstrates its ability to effectively identify relevant questions while maintaining a strong balance in correctly classifying rare negative samples.
In contrast, the TF-IDF-based models showed nearly identical performance across metrics, primarily predicting all samples as positive. Given the highly imbalanced label distribution (910 positive vs. 7 negative samples in the test set), these models achieve high recall by predicting positives but struggle to correctly identify the few negative instances, reflecting the challenge posed by the skewed data.

\subsubsection{Task 3: Answer Readability}
The evaluation results for the task \textit{answer readability} are shown in Table~\ref{eval3}. Among the ML-based models, the LR classifier exhibited the highest performance, achieving an F1 Micro score of 81.72\% and a QWK of 41.20\%. While LR provided reasonable performance, the relatively low F1 Macro score highlights its difficulties in handling class imbalance effectively. In contrast, the PLM-based models demonstrated significant improvements over the ML-based models. For instance, BERT achieved an F1 Micro score of 86.04\% and a QWK of 63.52\%, showcasing its superior ability to capture the complexities of readability assessment through richer contextual understanding. 
\paragraph{Detailed Evaluation of FinBERT's Performance.}
FinBERT outperformed all other models with an F1 Micro score of 87.57\%, an F1 Macro score of 41.34\%, an F1 Weighted score of 85.98\%, and the highest QWK of 66.53\%. Its pre-training on financial text provided a distinct advantage for this domain-specific task, enabling it to achieve the highest accuracy and alignment with human judgments. In further analysis, we find that the model performs well in recognizing answers with the highest readability, but it struggles to accurately differentiate between the lower readability levels, highlighting the need for improvement in capturing finer distinctions in readability.

\subsubsection{Task 4: Answer Relevance}
The evaluation results for the \textit{answer relevance} task are detailed in Table~\ref{eval4} for supervised models and Table~\ref{eval5} for zero-shot LLMs. As the most challenging criterion, assessing nuanced answer relevance revealed distinct performance variations among the different modeling strategies. Our key observations are summarized as follows:
\begin{table}[t]

  \centering
  \resizebox{1\linewidth}{!}{
  \begin{tabular}{llcccc}
    \toprule
   \textbf{Category} & \textbf{Model} & \textbf{F1 (Mi)}    &  \textbf{F1 (Ma)}  & \textbf{F1 (W)} & \textbf{QWK}\\
     \hline
     \multirow{4}*{ML-based}&TF-IDF + LR  &  69.61$\pm$0.49 & \textbf{49.74}$\pm$0.53 & \textbf{65.07}$\pm$0.60 & \textbf{38.16}$\pm$2.13\\
     
     &TF-IDF + RF  &  69.97$\pm$0.34 & 44.79$\pm$ 0.84&63.07$\pm$0.76 &36.63$\pm$1.10\\
     
     &TF-IDF + SVM  &  \textbf{70.08}$\pm$0.83 & 44.42$\pm$0.43 & 62.93$\pm$1.07 & 36.78$\pm$1.65\\
     
     &TF-IDF + KNN  &  66.66$\pm$1.03 & 38.63$\pm$2.05 &57.87$\pm$1.42 &23.22$\pm$3.36 \\
     
    \hline
    \multirow{9}*{PLM-based}&BERT &  77.90$\pm$0.98 & 67.64$\pm$0.94& 76.79$\pm$0.75 & 63.95$\pm$0.85\\ 
    


    &RoBERTa &  77.10$\pm$1.34 & 65.95$\pm$2.64 & 75.77$\pm$1.35 & 62.22$\pm$2.46\\ 

    &FinBERT &  78.19$\pm$0.96 & 68.18$\pm$0.79 & 77.25$\pm$0.81 & 65.49$\pm$0.54\\ 

    &Mengzi-fin &  78.04$\pm$0.72 & \textbf{68.96}$\pm$2.42 & \textbf{77.44}$\pm$0.82 & \textbf{66.31}$\pm$1.10\\ 

    &DKPLM &  76.92$\pm$0.57 & 63.00$\pm$ 3.72& 74.65$\pm$1.15& 62.82$\pm$1.78\\ 

    &SBERT-nli &  74.52$\pm$0.36 & 61.18$\pm$2.54 & 72.38$\pm$1.07 & 53.88$\pm$1.12\\ 

    &RoBERTa-extractive-qa &  \textbf{78.33}$\pm$0.36 & 67.02$\pm$1.48 &77.12$\pm$0.69 & 64.66$\pm$1.44\\ 

    &BERT (Large) &  78.01$\pm$0.78 & 65.94$\pm$1.62 & 76.16$\pm$0.56 & 63.53$\pm$0.90\\ 
    &RoBERTa (Large) &  78.30$\pm$0.94 & 68.50$\pm$2.09 & 77.40$\pm$1.54 & 64.52$\pm$1.56\\  
    \hline
    \multirow{4}*{PLM with CPT-based}&FinBERT &  \textbf{80.04}$\pm$0.23 & \textbf{70.58}$\pm$0.90 & \textbf{79.02}$\pm$0.18 & \textbf{67.89}$\pm$1.26\\ 
    &RoBERTa-extractive-qa &  78.34$\pm$0.34 & 68.01$\pm$ 1.71& 77.45$\pm$0.74 & 66.76$\pm$0.58\\ 
    &BERT (Large) & 78.12 $\pm$0.65 &65.15$\pm$3.97 & 76.02$\pm$ 1.16& 64.82$\pm$2.07\\  
    &RoBERTa (Large) &  74.37$\pm$1.78 & 61.40$\pm$1.28 & 72.46$\pm$0.73 & 53.91$\pm$3.42\\

    \bottomrule
  \end{tabular}
}

  \caption{Model evaluation results on the \textit{answer relevance} task. 
  }
  \label{eval4}

\end{table}

\begin{table}

  \centering
  \resizebox{1\linewidth}{!}{
  \begin{tabular}{lcccccccc}
    \toprule
\multirow{2}*{\textbf{Model}} &
\multicolumn{4}{c}{\textbf{Direct}} &
\multicolumn{4}{c}{\textbf{CoT}} \\
\cmidrule(lr){2-5}\cmidrule(lr){6-9}
& \textbf{F1(Mi)}  &  \textbf{F1(Ma)}  & \textbf{F1(W)} & \textbf{QWK}  & \textbf{F1(Mi)}  &  \textbf{F1(Ma)}  & \textbf{F1(W)} & \textbf{QWK}\\
    \hline
    GPT-4  &  \textbf{67.39} &\textbf{63.11} &\textbf{70.14} & \textbf{59.06} &  59.76 & 58.45&64.28&57.51\\

    GPT-4o &  54.53 & 53.02 & 59.76  & 53.34 & 54.96  & 54.73 & 58.25 & 47.97 \\

    GPT-5 &  60.52 & 59.45 & 62.66  & 51.91 & 59.32  & 58.32 & 61.10 & 49.86 \\

    O3 &  65.43 & 62.88 & 68.37  & 57.93 & \textbf{66.41}  & \textbf{64.29} & \textbf{69.24} & \textbf{60.12} \\

    GPT-OSS 120B & 62.27 & 60.55 & 65.35 & 55.76 & 61.94 & 59.94 & 64.81 & 53.87 \\

    Llama-3.1 8B Instruct  & 53.00&  41.59 & 56.13&43.24 &  46.46 &40.82 &51.57&39.56\\

    Qwen2 7B Instruct & 60.52&  31.33 & 52.83 & 23.68 &  59.00 & 38.97&57.36&42.35\\
    
    Mistral 7B Instruct &  54.53 & 30.40 & 50.99 & 29.57&  59.32 & 23.03&60.20&38.90\\
    
    Qwen3-Next 80B Instruct & 59.98 & 56.71 & 64.51 & 56.23 & 60.09 & 57.09 & 61.38 & 43.20 \\

    Qwen3-Next 80B Thinking & 59.11 & 56.26 & 60.50 & 44.83 & 58.89 & 56.62 & 59.82 & 42.98 \\
    
    \bottomrule
  \end{tabular}
}

\caption{Zero-shot evaluation results of Large Language Models (LLMs) on the \textit{answer relevance} task, comparing direct scoring against chain-of-thought (CoT) prompting.}
  \label{eval5}
\end{table}

\paragraph{PLM-based Methods Consistently Outperform ML-based ones. }
As shown in Table~\ref{eval4}, all PLM-based methods can achieve better performance than ML-based methods across all metrics. For instance, RoBERTa-extractive-qa achieved the highest Micro F1 score of 78.33\%, while Mengzi-fin excelled with the highest Macro F1 score of 68.96\% and a strong QWK of 66.31\%. Their superior performance can be attributed to pre-training on task-specific and domain-specific data: RoBERTa-extractive-qa was pre-trained on QA datasets, which enhanced its ability to comprehend and evaluate answer relevance, while Mengzi-fin was pre-trained on financial corpora, enabling it to better capture domain-specific nuances in the financial context.

\paragraph{Impact of Continued Pre-training on PLM-based Methods.}
We selected four PLM-based models and further pre-trained them using 900,000 Q\&A pairs. As shown in Table~\ref{eval4}, we observed that only FinBERT demonstrated significant performance improvements, achieving the highest Micro F1 score of 80.04\% and a QWK of 67.89\%, indicating its enhanced ability to assess answer relevance in financial contexts. In contrast, the other three models (RoBERTa-extractive-qa, BERT Large, and RoBERTa Large) did not exhibit consistent improvements. This suggests that domain alignment plays a crucial role in the effectiveness of continued pre-training, as FinBERT’s domain-specific pre-training on financial texts provided a strong foundation, whereas the other models may have experienced a mismatch between their learned representations and the specific requirements of the task, resulting in less notable gains.

\paragraph{LLMs Still Struggle with Fine-Grained Relevance Judgments in Chinese Financial QA.}
As shown in Table~\ref{eval5}, we evaluated advanced LLMs on the answer relevance task under both direct scoring and CoT prompting. GPT-4 achieves the best performance in the direct setting, while O3 performs best under CoT. However, the effect of CoT is highly model-dependent: it substantially improves O3, but does not benefit GPT-4 or GPT-5. This suggests that explicit reasoning is not universally helpful for fine-grained relevance assessment in Chinese financial disclosures. More broadly, even the strongest LLMs remain behind the best supervised PLM-based models, indicating that this task remains challenging for general-purpose LLMs. We attribute this difficulty to the need for domain-specific financial knowledge, the subtle distinction between adjacent relevance labels, and the prevalence of indirect or evasive disclosure patterns in real investor--company interactions.

We also experimented with Chinese financial LLMs, including CFGPT2~\citep{li2023cfgpt} and DISC-FinLLM~\citep{chen2023disc}. However, these models frequently failed to follow the scoring instructions and often produced repetitive or malformed outputs, making their predictions unreliable. We therefore exclude them from the main comparison.

\begin{figure}[t]
  \centering
  \begin{subfigure}[t]{0.48\linewidth}
      \centering
      \includegraphics[width=\linewidth]{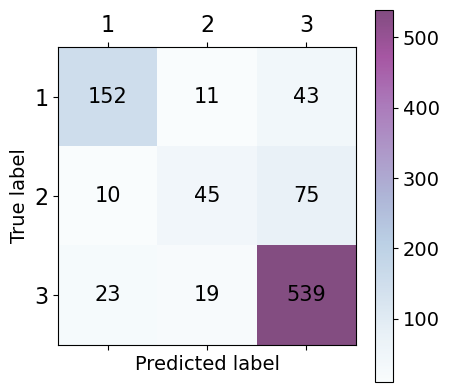}
      \subcaption{FinBERT with continued pretraining for \textit{answer relevance} task.}
      \label{conf4}
  \end{subfigure}
  \hfill 
  \begin{subfigure}[t]{0.48\linewidth}
      \centering
      \includegraphics[width=\linewidth]{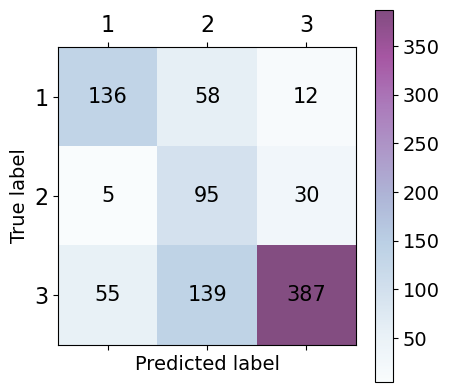}
      \subcaption{GPT-4 (direct score) for \textit{answer relevance} task.}
      \label{gptconf}
  \end{subfigure}

  \caption{Confusion matrices for \textit{answer relevance} predictions from FinBERT with continual pretraining and GPT-4 (direct scoring). 
  }
\end{figure}

\paragraph{Error Analysis.}
To better understand the gap between supervised PLMs and LLMs, we further analyze their confusion patterns. Figure~\ref{conf4} presents the confusion matrix of the best-performing supervised model, FinBERT with continued pretraining. The model reliably identifies answers that are fully relevant to the question (label 3), but often confuses partially irrelevant answers (label 2) with fully relevant ones, suggesting that the boundary between these two categories is especially subtle. By contrast, the confusion matrix of GPT-4 shown in Figure~\ref{gptconf} reveals more frequent confusion between completely irrelevant answers (label 1) and partially relevant ones (label 2), together with a tendency to over-predict the middle category. This pattern indicates weaker calibration in borderline cases, especially when an answer contains superficially related financial terminology without directly addressing the question. Additional qualitative analysis of representative failure cases can be found in our GitHub repository.
\section{Conclusion}
This study introduces FinTruthQA, a benchmark designed to evaluate advanced NLP techniques for the automatic quality assessment of information disclosure in financial Q\&A data. We constructed a dataset of 6,000 real-world financial Q\&A entries, each manually annotated based on four key accounting-oriented evaluation criteria. Experiments revealed the effectiveness of pre-trained BERT models in tackling these tasks. In particular, we observed that models pre-trained on financial corpora, as well as those designed for related tasks such as QA, outperformed other variants of BERT and GPT-4. Although existing NLP models showed reasonable performance on this benchmark, there is still significant room for improvement, especially in the task of \textit{answer relevance}, which calls for more robust and business-aware NLP models. Beyond academic evaluation, FinTruthQA provides a practical foundation for AI-driven disclosure monitoring in capital markets, supporting disclosure governance, investor-facing communication, and data-driven decision-making in real-world financial settings. Our benchmark is useful not only for post-hoc regulatory assessment but also for real-time applications that promote financial transparency, reduce information asymmetry, and help auditors, financial analysts, regulators, and corporate communication teams identify inadequate or potentially misleading responses in a timely manner.

\paragraph{Limitations and Future Work.} The Chinese cultural and regulatory context of SSE and SZSE may limit generalizability to other markets, and the manual annotation process inherently involves a degree of subjectivity. Furthermore, reliance on standard quantitative metrics may neglect interpretability, the \textit{question relevance} dataset is highly imbalanced (though reflective of real-world distributions), and computational constraints restricted our LLM evaluations to zero-shot settings. Future work will explore fine-tuned LLMs, retrieval-augmented methods, and human-centered evaluation protocols, while also validating these methodologies across diverse global regulatory and business contexts.
\bibliography{aaai2025-condensed}
\end{document}